\documentclass[
twocolumn,
]{ceurart}

\usepackage{listings}
\usepackage{multirow}
\usepackage{graphicx}
\usepackage{enumerate}
\usepackage[shortlabels]{enumitem}
\usepackage{multirow}
\usepackage{hyperref}
\begin{document}

\copyrightyear{2024}
\copyrightclause{Copyright for this paper by its authors.
  Use permitted under Creative Commons License Attribution 4.0
  International (CC BY 4.0).}

\conference{Ital-IA 2024: 4th National Conference on Artificial Intelligence, organized by CINI, May 29-30, 2024, Naples, Italy}

\title{Graph Neural Networks for Gut Microbiome Metaomic data: A preliminary work}

% \author[1]{Christopher Irwin}[%
% email=christopher.irwin@uniupo.it,
% ]
% \cormark[1]
% \address[1]{DISIT, Computer Science Institute,
% University of Piemonte Orientale, Alessandria, Italy}

\author[1,3]{Christopher Irwin}[%
email=christopher.irwin@uniupo.it,
]
\cormark[1]
\fnmark[2]
\author[2]{Flavio Mignone}[%
email=flavio.mignone@smartseq.it,
]
\author[1,3]{Stefania Montani}[%
email=stefania.montani@uniupo.it,
]
\author[1,3]{Luigi Portinale}[%
email=luigi.portinale@uniupo.it,
]

\address[1]{DISIT, Computer Science Institute,
University of Piemonte Orientale, Alessandria, Italy}
\address[2]{SmartSeq s.r.l., Alessandria, Italy}
\address[3]{Interdepartmental Research Center on Artificial Intelligence (AI@UPO), University of Piemonte Orientale, 15121 Alessandria, Italy}

\cortext[1]{Corresponding author.}
\fntext[1]{Authors are listed in alphabetical order.}
\fntext[2]{Phd student enrolled in the National PhD in Artificial Intelligence for Health and Life Sciences, XXXVIII cycle, Università Campus Bio-Medico, Roma.}

\begin{abstract}
The gut microbiome, crucial for human health, presents challenges in analyzing its complex metaomic data due to high dimensionality and sparsity. Traditional methods struggle to capture its intricate relationships. We investigate graph neural networks (GNNs) for this task, aiming to derive meaningful representations of individual gut microbiomes. Unlike methods relying solely on taxa abundance, we directly leverage phylogenetic relationships, in order to obtain a generalized encoder for taxa networks. The representation learnt from the encoder are then used to train a model for phenotype prediction such as Inflammatory Bowel Disease (IBD).
\end{abstract}


\maketitle

\section{Introduction}
The gut microbiome, a community of microorganisms residing within the human gut, plays a significant role in health and disease. Advancements in sequencing have yielded vast amounts of metaomic data, detailing the gut microbiome's composition and function. However, effectively analyzing this high-dimensional data remains a challenge. Traditional methods often fail to capture the intricate relationships between the diverse microbial species within the microbiome.

This work explores the potential of graph neural networks (GNNs) for analyzing gut microbiome metaomic data. GNNs, a powerful class of deep learning models adept at handling graph-structured data, hold promise for unraveling the complex relationships within the microbiome. Our primary objective is to develop a method that utilizes metaomic data to obtain a meaningful representation of a patient's gut microbiome. In particular, we begin by constructing a network that captures the relationships between genes, species, and genus. This network is built using the gene expression levels obtained from all patient samples. By leveraging GNN techniques, we then learn latent representations (embeddings) for each entity within the network. These embeddings capture the underlying functional relationships between genes, species, and genus. Finally, we aggregate these embeddings based on a patient's specific gene expression profile to obtain a global representation of their unique gut microbiome. This patient-specific microbiome representation can then be used by a classifier to predict a particular phenotype, such as the presence or absence of Inflammatory Bowel Disease (IBD) as done in the present work.

%\noindent
Recently, a wide rage of methods have been proposed to analyse microbiome data and use it to obtain predictive models for biomarkers discovery and classifying patients based on some phenotype. Some of these methods rely on the use of the microbiome features like the relative abundance of taxa. However, such approaches can be limited by the inherent nature of microbiome datasets: sparsity and high dimensionality (e.g., the presence of many features). This may lead to overfitting, with models failing to generalize to other datasets~\cite{miostone}.

In order to address these kind of limitations, many works try to leverage the intrinsic phylogenetic information between taxa \cite{reviewnat}. For example PopPhy-CNN \cite{popphycnn} uses a CNN-based model that works on the 2D matrix derived from the phylogenetic tree representation. MIOSTONE \cite{miostone} uses a gated phylogenetic encoded neural network. TaxoNN \cite{taxonn} leverages an ensemble of CNNs each specializing to particular taxonomic phylum. \newline

\noindent
Our method directly builds upon the concept of leveraging phylogenetic relationships to create a meaningful representation of the microbiome. We achieve this by employing a GNN-based method to learn a generalized encoder specifically for the network of taxa. This encoder is trained independently from the downstream classification task, which is handled by a separate model.

\section{Dataset} \label{sec:dataset}
The dataset in use is the IBDMDB which is described in \cite{idbmdb} related to a IBD cohort of patients. The database contains observations coming from 2 different omic levels: microbiome metagenomics (MGX) and metatranscriptomics (MTX) gene expression levels (expressed in count per milion - CPMs) along with some other information related to the patient state and the presence of absence of IBD.

Different omic levels have different number of samples and different number of features, and in particular:

\begin{itemize}
    \item \textbf{Metagenomics}: 1635 samples with 108433 features.
    \item \textbf{Metatranscriptomics}: 736 samples with 70711 features.
\end{itemize}

Regarding the patient labels, these are related to the presence of absence of IDB (binary). Notably, there are 1594 patients with labels. The dataset has a quite substantial skew, since there are 83\% negative and 17\% positive samples.%\newline

% \noindent
The dataset has a large dimensionality and a different rate of missing samples. To this end, our approach is  to use methods which can deal with missing samples for a particular omic-level and a different number of features. Moreover, as already expressed in the previous section, when dealing with metaomics, it can potentially be useful to being able to model an interaction between the different microorganisms.

\begin{figure*}[ht]
    \centering
    \includegraphics[width=0.8\linewidth]{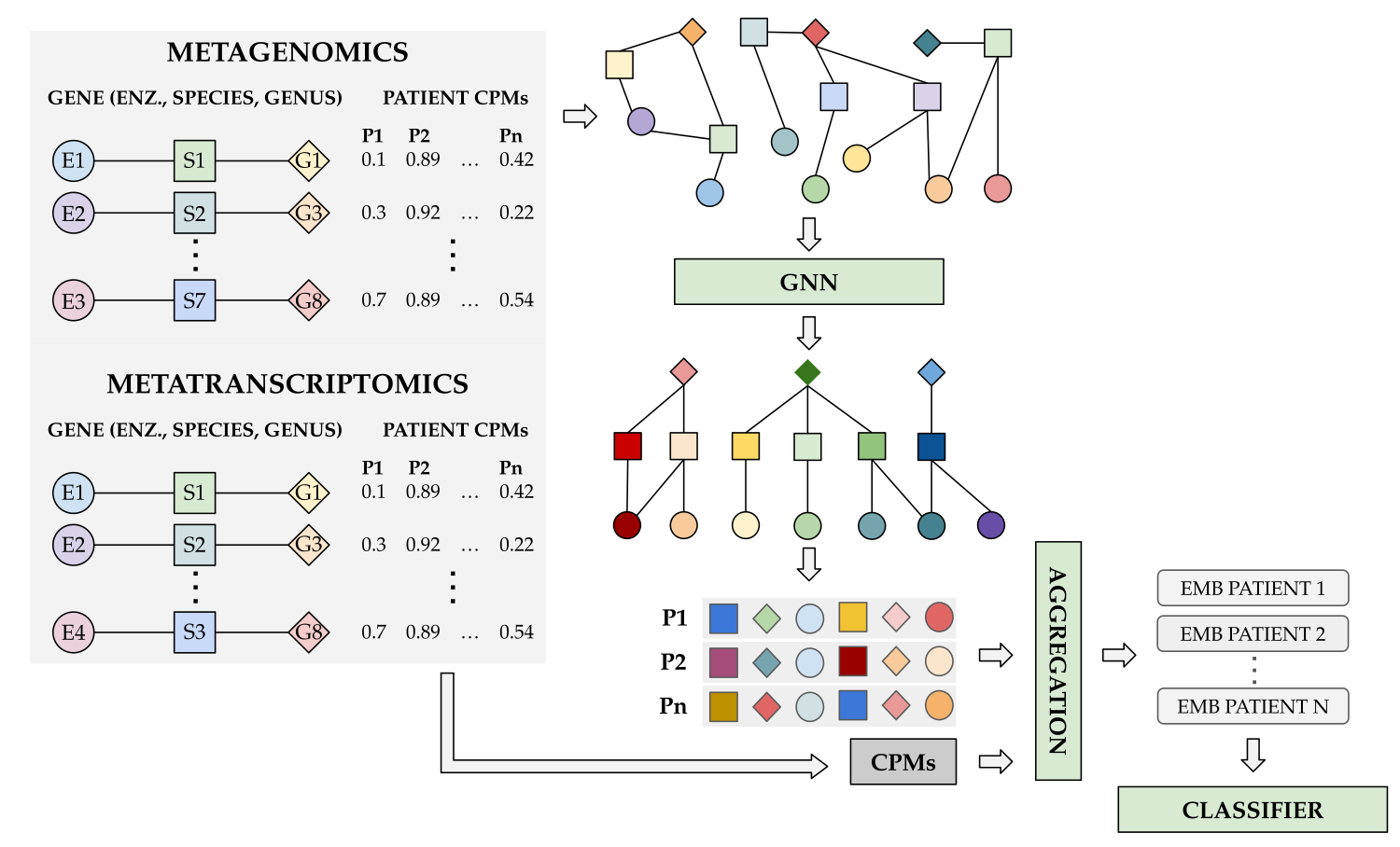}
    \caption{Overview of the proposed method: input data characterized by microbiome gene expression across different omic levels (left). Extracted graph connecting genes based on species and genus (center). Application of the GNN module to contextualize node representations: the set of node embeddings representing a patient are aggregated by taking in to account the relative abundance expressed in CPMs and fed to the final classifier predicting IBD presence (bottom)}
    \label{architecture}
\end{figure*}

\section{Proposed Method}
The main task we want to carry out is to obtain a meaningful representation of a patient using the meta-omic data coming from the analysis of gut microbiome samples.
To do this, the idea is to establish relationships among the various microorganisms present in the microbiome, whose presence is determined by the abundance of enzymes/genes that characterize them, using the relational information of species and genus encoded within a graph.
Moreover, since it is common to have incomplete datasets (either patients may have different gene expression or missing data may come from a particular omic level), we model our task and the way of treating the data in a way that allows transparency from this point of view. Figure~\ref{architecture} gives an overview of the proposed pipeline. %\newline

%\noindent
The following subsections will describe how to get the representation of a patient starting from the metagenomic level. The same technique can be applied in the same way to other omic levels and seamlessly integrated in this pipeline.

\subsection{Graph Construction}
Given a set of patients with gene expression levels of gut microbiome microorganisms, and associated data on enzymes, species, and genus (phylogenetic information) for each microorganism, our aim is to construct a graph $\mathcal{G}=(U,V)$ where:
\begin{itemize}
    \item $U$ denotes the set of nodes, encompassing enzymes, species, and genus.
    \item $V$ represents the set of edges, categorized into two types: \texttt{(enzyme, species)} and \texttt{(species, genus)}.
\end{itemize}

Once the phylogenetic graph is constructed, we can represent each patient $P_i$ as a subset of nodes $U_i \subseteq U$, where the patient exhibits a gene expression level greater than 0.

\subsection{Graph Representation Learning Module}
Given a graph $\mathcal{G}$ that encodes the relationships between genes, we aim to learn a representation (embedding) of the nodes that reflects their connectivity within the network. To achieve this, we leverage established methods from the field of graph representation learning, specifically: Graph Laplacian Eigenvector Positional Encoding, Random Walk Positional Encoding, and Node2Vec. %\newline

\noindent
Here, we provide a concise description of each method. \newline

\noindent
\textbf{Graph Laplacian Eigenvector Positional Encoding (LPE)}: this technique relies on the factorization of the graph's Laplacian  matrix. The embedding for a node is defined by the k-smallest,  non-trivial eigenvectors associated with that node \cite{posenc}. \newline

\noindent
\textbf{Random Walk Positional Encoding (RWPE)}: This approach, as described in \cite{posenc}, utilizes the random walk matrix to generate node embeddings. \newline

\noindent
\textbf{Node2Vec (N2V)}: This method, presented in \cite{n2v}, is a semi-supervised algorithm that leverages  random walks on the graph to learn meaningful features for the nodes.

\subsection{Aggregation Function}
Having obtained embeddings for all the nodes in the graph, we now require a methodology to aggregate a subset of these embeddings into a single vector that represents a patient. Two aggregation levels are employed for this purpose. 

The first level of aggregation aims to generate a flattened representation for each gene. Here, a gene's embedding is calculated as the mean of the embeddings associated with its corresponding phylogenetic subgraph. 

The second level of aggregation focuses on converting the set of genes expressed by a patient into a single embedding representation. This is achieved by utilizing a hyperparameter that determines the number of most highly expressed genes considered for the aggregation process. The optimal value of this hyperparameter will be investigated in Section \ref{sec:q3}.

\subsection{Different Omic Levels Integration}
Previous sections have outlined the methodology for transforming gene abundance data (obtained from a metagenomic analysis) of a patient's microbiome into a single patient representation. However, incorporating metatranscriptomic features is straightforward. We simply include genes identified at the transcript level during the construction of the phylogenetic graph, these additional nodes are then considered when generating the patient representation.

\section{Experimental results}
To evaluate the performance of our proposed method, we conducted preliminary experiments using the IBD dataset described in Section \ref{sec:dataset}. These experiments aimed to investigate the method's effectiveness in predicting the presence or absence of IBD under various parameter settings. We addressed the following key questions:

\begin{enumerate}[start=1,label={\bfseries Q\arabic*.}]
    \item \textbf{Node Embedding Comparison:} which node embedding technique (LPE, RWPE, N2V) yields the best performance?
    \item \textbf{Impact of Multi-Omics Integration:} how does the model's performance vary when using different omic levels  as input: metagenomics only versus metagenomics combined with  metatranscriptomics?
    \item \textbf{Gene Selection Strategy:} how does the number of genes considered during patient representation generation impact the final model performance?
    
\end{enumerate}

For the final classification task utilizing patient representations as input features, we employed a Support Vector Machine (SVM) classifier using a Radial Basis Function (RBF). 
Our choice of a relatively simple model reflects our primary focus on evaluating the efficacy of the graph representation techniques.

\subsection{Evaluation of Node Embedding Techniques (Q1)}
To assess the performance of different node embedding techniques (LPE, RWPE, N2V), we employed an 80/10/10 train-validation-test split. For each technique, we optimized model hyperparameters using the Optuna framework \cite{optuna}. Following hyperparameter tuning, each optimal model configuration was evaluated over independent runs, utilizing randomized train-validation-test splits and random initialization. %\newline
Table~\ref{tab:resq1} reports the results related to Q1. It is clear from such results that the best choices are LPE and N2V, which perform very similarly on both datasets. While RWPE significantly underperforms with repsect to other methods.

\subsection{Impact of Omic Levels (Q2)}
To assess the influence of data derived from different omic levels on  model performance (Q2), we trained two distinct models. The first model used only data from the MGX level, while the second incorporated data  from both metagenomics (MGX) and metatrascriptomics (MTX) levels. Importantly, the  evaluation for Q2 employed the same experimental setup established for Q1 (detailed in the previous section). The corresponding results for both models are presented in Table~\ref{tab:resq1}. From the results it is possible to notice that the model trained using only MGX data sligthly underperforms with respect to the model that includes MTX data.

\begin{table}[h]
\centering
\caption{Results obtained by the models using different methods for encoding the phylogenetic graph}
\label{tab:resq1}
\resizebox{\columnwidth}{!}{%
\begin{tabular}{l|ccc|ccc}
\multicolumn{1}{c|}{\multirow{2}{*}{\textbf{Dataset}}} & \multicolumn{3}{c|}{\textbf{F1 score}}      & \multicolumn{3}{c}{\textbf{ROC AUC}}        \\
\multicolumn{1}{c|}{}                         & \textbf{LPE} & \textbf{RWPE} & \textbf{N2V} & \textbf{LPE} & \textbf{RWPE} & \textbf{N2V} \\ \hline
MGX only & 0.839 & 0.664 & 0.837 & 0.908 & 0.709 & 0.914 \\
MGX+MTX  & 0.871 & 0.681 & 0.873 & 0.929 & 0.721 & 0.923
\end{tabular}%
}
\label{tab:resq1}
\end{table}

\begin{figure*}[h]
    \centering
    \includegraphics[width=0.9\linewidth]{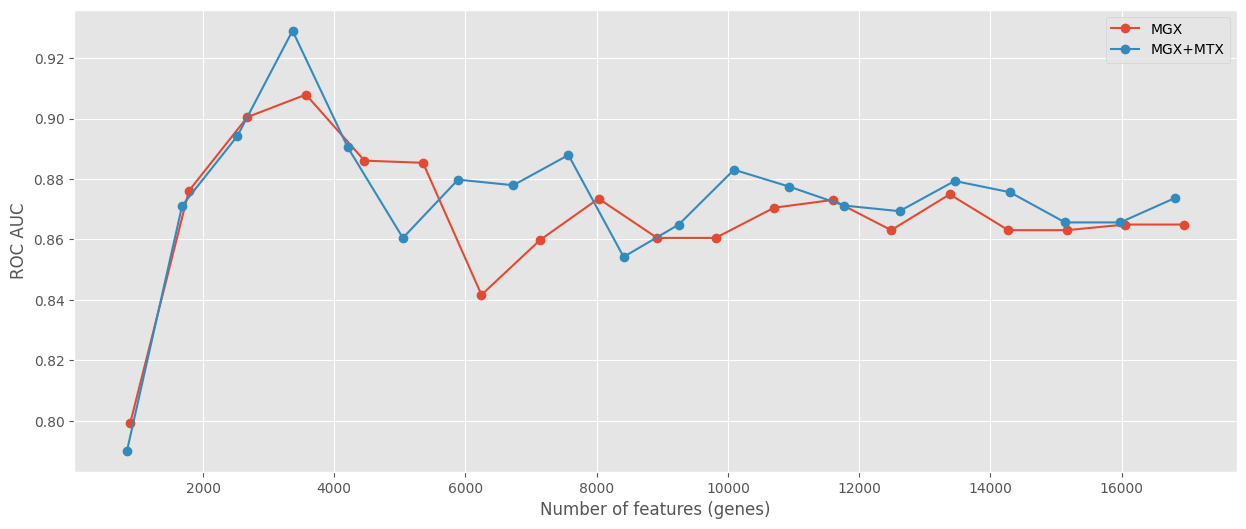}
    \caption{The plot compares the ROC AUC achieved by the model trained on two different datasets: MGX and MGX+MTX. In this case the graph encoder used the LPE tecnhique for the node embedding representation.}
    \label{q3_res}
\end{figure*}

\subsection{Gene Selection Strategy (Q3)} \label{sec:q3}
Relatively to Q3, we focused on the model trained using the LPE encoding for the phylogenetic graph. To obtain the representation of a patient, we trained the model using as features a different number $k$ of genes. In particular, we
considered the top-$k$ most expressed genes. Moreover, we also compared the results between the two datasets (MGX-only, MGX+MTX). As shown in Figure~\ref{q3_res}, the model's ROC-AUC initially increases with the number of features, peaking at a certain point before a slight decrease and stabilization. This likely reflects the importance of highly expressed genes for class separation. Genes with lower expression levels are likely to carry less information. Interestingly, despite a small difference, the optimal number of genes differs between the datasets. The MGX-only model achieves its peak ROC-AUC with 3568 genes, while the MGX+MTX model peaks at 3366. This might suggest that the MTX data provides additional useful information, resulting in a smaller number of necessary genes.

\section{Conclusion and future work}
This paper proposes a method leveraging Graph Neural Networks (GNNs) to extract phylogenetic information from microbiome metagenomics data. The method aims to create a generalizable encoder for the microbiome, enabling its application to phenotype classification tasks. The approach achieves good performance on real-world datasets even with simple classification models, although it falls short of models specifically designed for classification tasks~\cite{imovnn}. Future work will involve:
\begin{itemize}
    \item Expanding comparisons with other methods.
    \item Evaluating performance on additional datasets.
    \item Exploring the use of clustering for phenotype classification using the learned encoder.
\end{itemize}

\begin{acknowledgments}
We acknowledge the use of the Chameleon Cloud platform~\cite{chameleon} for providing the essential computational resources that facilitated the execution of our experiments and model training. 
% Christopher Irwin is a PhD student enrolled in the National PhD in Artificial Intelligence, XXXVIII cycle, course on Health and life sciences, organized by Università Campus Bio-Medico di Roma.
\end{acknowledgments}

%%
%% Define the bibliography file to be used
\bibliography{sample-ceur}

\end{document}